\documentclass[letterpaper,10pt,conference]{ieeeconf}

\usepackage{graphicx}
\usepackage{amsmath}

\title{Vehicle-State-Aware Confidence Estimation for Direct Wind Sensing on a Multirotor}

\author{Sri Arjun Senthil Kumar, April Zi Qian Chan, Chen-Hua Yeow}

\begin{document}
\maketitle

%
%
%


\begin{abstract}
Direct wind sensing from a multi-rotor UAV enables localized atmospheric measurements, but the reliability of the anemometer depends on vehicle motion. This work presents a vehicle-state-aware confidence method that uses horizontal and vertical vehicle speed to quantify the acquisition of each wind observation. The resulting confidence score is used to adapt measurement influence to a two-dimensional vector estimator. Experiments were conducted during horizontal translation, vertical motion, and held-out autonomous waypoint flight. The proposed two-feature formulation achieved an  ROC--AUC of 0.995, with 98.84\% of translation
observations identified as unreliable. The confidence-aware adaptation reduced median estimator gain during translation from 0.207 to 0.064 while retaining median gain of 0.151 during hover. The results show that vehicle state provides useful proprioceptive context for regulating the influence of direct airborne wind observations.

\end{abstract}


\section{Introduction}

Unmanned aerial vehicles provide flexible platforms for localized atmospheric observation as they can hover at required locations and perform vertical hovers and traverse short trajectories in the atmospheric boundary layer ~\cite{palomaki2017wind,wetz2021distributed}. Multirotor platforms are suitable for in-situ wind, sensing as a lightweight sonic anemometer can be carried to locations difficult to instrument using fixed towers ~\cite{thielicke2021accurate,ghirardelli2025samurai}. 

Direct airborne wind sensing introduces a coupling between environmental observation and the motion of the sensing platform, as the anemometer is exposed not only to atmospheric flow but also to changes in relative airflow associated with vehicle translation and the local aerodynamic environment generated by the propulsion system. Sensor elevation can reduce these effects but does not eliminate them ~\cite{thielicke2021accurate,
yang2025correction,ghirardelli2025samurai}. Additionally, the calibration of the anemometer alone does not imply that every airborne wind observation is under favorable sensing conditions. 

The vehicle motion itself provides information about the acquisition condition. Previous work has used attitude, internal measurements, motor information and learned dynamic models to estimate wind without a dedicated sensor ~\cite{neumann2015realtime,gonzalezrocha2019sensing,
wetz2021distributed,zimmerman2022wind,chen2026wind}. These approaches show the relationship between wind and multirotor dynamics. However, vehicle state is primarily used for reconstructing wind fields.

This work takes the approach where rather than estimating ambient wind directly from vehicle dynamics, we use the instantaneous vehicle state to quantify vehicle conditions associated with \emph{direct} onboard wind observation. Horizontal and absolute vertical speeds are compared with an empirically determined stable hover reference to obtain a continuous wind observation confidence score. The score is used to adapt the measurement covariance of a navigation-frame wind-vector estimator, reducing the influence of observations obtained during strong departure from stable sensing while preserving measurement streams. 

The contributions of this work are:

\begin{itemize}
    \item A proprioceptive confidence model that uses multirotor motion state to assign sample-level confidence to direct wind observations, enabling motion-aware adaptation of measurement uncertainty in downstream state estimation without discarding observations or requiring a vehicle dynamic model. 

    \item A confidence-aware vector wind estimator that propagates
    calibrated wind-speed and direction uncertainty into a
    sample-dependent measurement covariance and continuously adapts
    measurement influence using the vehicle-state confidence score.

    \item Experimental validation across development, independent vertical-motion, and held-out autonomous waypoint flights, including feature ablation and estimator sensitivity analyses, demonstrates that the frozen confidence models transfer across distinct flight regimes and selectively reduce the influence of wind observation acquired under motion-disturbed sensing conditions.

\end{itemize}

\begin{figure*}[t]
    \centering
    \includegraphics[width=0.98\textwidth]
    {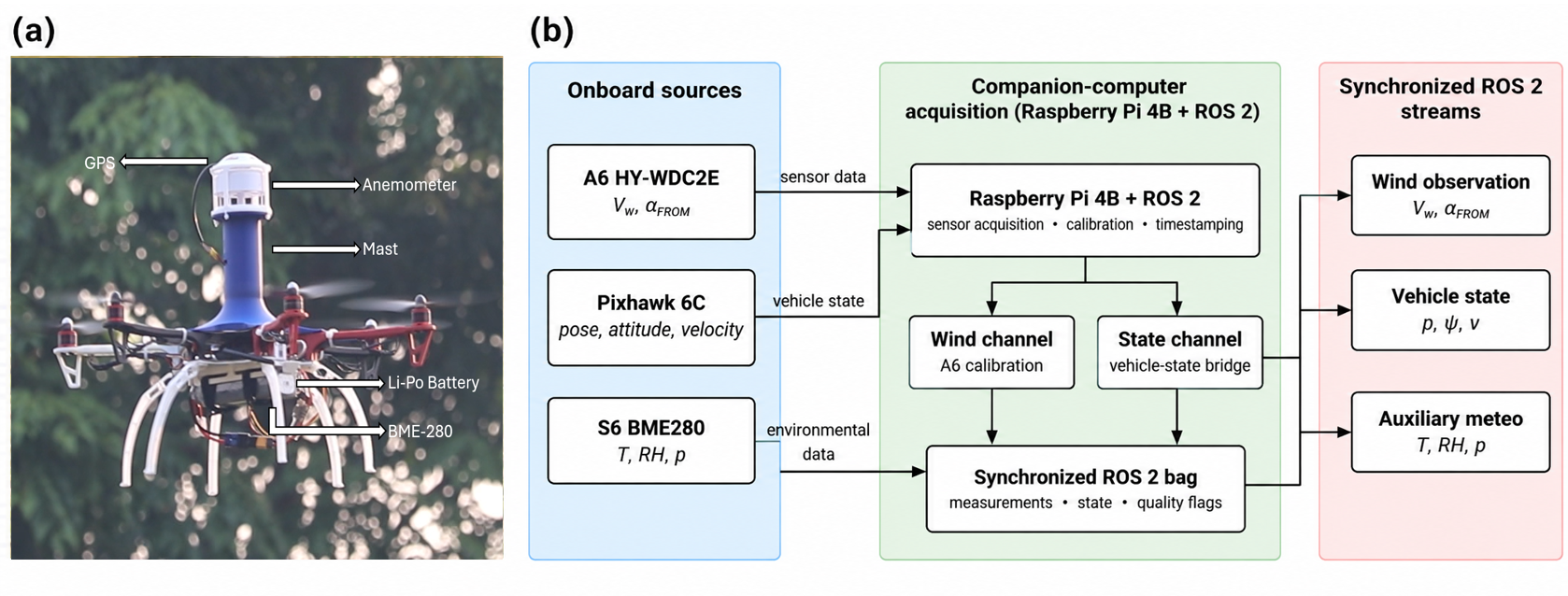}

    \caption{
    Representative D6 MeteoNode platform and onboard sensing architecture.
    (a) Outdoor experimental platform equipped with an elevated
    HY-WDC2E ultrasonic anemometer.
    (b) Environmental and vehicle-state measurements are acquired through
    the Raspberry Pi companion computer and Pixhawk flight controller and
    recorded through ROS~2 for downstream confidence-aware wind processing.
    }
    \label{fig:platform_architecture}
\end{figure*}

\section{Related Work}

\subsection{UAV-Based Atmospheric Wind Sensing}
Multirotor UAVs are increasingly used for atmospheric wind observation. Palomaki \emph{et al.} demonstrated direct wind sensing using a sonic anemometer above a hexacopter and indirect estimation using vehicle attitude ~\cite{palomaki2017wind}.
Wetz \emph{et al.} demonstrated distributed atmospheric boundary layer measurement using multiple quadrotors and a dynamic-based wind estimator method ~\cite{wetz2021distributed}. Recently, sonic anemometer systems have been developed for high-fidelity atmospheric measurements, including rigidly mounted configurations~\cite{thielicke2021accurate} and suspended sensor systems designed to further separate instrument from propulsion-induced flow~\cite{ghirardelli2025samurai}. These studies establish that multi-rotors are useful platforms and cannot be treated as aerodynamically independent measurements. 

\subsection{Direct Wind Measurement and Vehicle-Induced Disturbance}

Sensor placement and vehicle-induced flow are important considerations. Palomaki \emph{et al.} stated that measured wind speed increases for a sonic anemometer mounted above a hovering multi-rotor ~\cite{palomaki2017wind}. Thielicke \emph{et al.}
showed that the accurate multi-rotor wind measurement is dependent on anemometer separation from the propellers, with appropriate treatment for propeller-induced flow 

~\cite{thielicke2021accurate}. Yang \emph{et al.} developed a correction that accounts for rotor-induced airflow and changes in flight attitude during three-dimensional rotary UAV wind measurements ~\cite{yang2025correction}.

While these approaches seek to reduce or correct the aerodynamic disturbance, in the present work elevated mounting and calibration are retained but no aerodynamic correction is assumed. Instead the vehicle state is used to describe how far instantaneous acquisition condition departs from stable-hover phase. 

\subsection{Vehicle-State-Based Wind Estimation and Sensing Confidence}

Other methods estimate wind from the vehicle itself. Neumann and Bartholmai estimated wind data using an onboard inertial measurement unit~\cite{neumann2015realtime}, González-Rocha \emph{et al.} evaluated kinematic and dynamic motion models to estimate wind from quadrotor motion~\cite{gonzalezrocha2019sensing}, while Wetz \emph{et al.} estimated based on aerodynamic-force-based formulation for atmospheric measurement ~\cite{wetz2021distributed}. Data-driven alternatives have also been investigated. Zimmerman\emph{et al.} used machine-learning models using multirotor dynamic and motor information ~\cite{zimmerman2022wind}. Work by Chen \emph{et al.} shows that the relationship between UAV attitude and wind depends on relative wind direction and platform configuration ~\cite{chen2026wind}.

The anemometer observation remains the wind measurement while the vehicle state supplies proprioceptive context describing the acquisition condition. This allows a sample-level confidence score to be assigned for direct observation and subsequently used to adapt its influence in the wind-vector estimator. Unlike dynamics-base approaches that reconstruct wind from vehicle response, the proposed method retains the anemometer as the primary wind sensor and uses proprioceptive vehicle state to assign confidence to each observation based on its acquisition condition. This addresses a complementary robotics problem: how the robot's own motion state can inform when an environmental measurement should be trusted and how strongly it should influence downstream estimation.


\section{Multirotor Sensing Platform and Characterization}
\label{sec:platform}

\subsection{Representative D6 Sensing Platform}

The platform used in this work is a DJI F550 hexacopter frame equipped with Pixhawk 6C flight controller and a Raspberry Pi 4B as the companion computer. The MeteoNode payload has a BME280 environmental sensor and a HY-WDC2E ultrasonic anemometer to observe wind speed and direction measurements. The anemometer is mounted in the center of the frame on an elevated mast to reduce the effects of propeller wash. Although the placement reduces direct aerodynamic interference from propeller wash, it does not fully isolate the sensor from vehicle-induced flow. 

\subsection{Onboard Wind and Vehicle-State Acquisition}

The Raspberry Pi acquires the wind measurement, applies sensor-specific calibration and  publishes the resulting observation through ROS~2. Vehicle position, attitude, and linear velocity are obtained from the Pixhawk through  the onboard MAVLink interface. The wind and vehicle state are recorded together for synchronized processing. 

Direct wind observations are associated with the nearest state sample within a temporal tolerance; unmatched observations are treated as unavailable. 

\subsection{A6 Wind-Sensor Characterization}
The wind speed calibration was performed using an Extech 45158 as the reference anemometer over $1.55$--$3.50~\mathrm{m\,s^{-1}}$, and the resulting A6 calibration was:
\begin{equation}
V_{\mathrm{cal}}
=
1.017 V_{\mathrm{raw}}-0.170.
\label{eq:a6cal}
\end{equation}

The RMSE during the calibration was nearly $0.069~\mathrm{m\,s^{-1}}$. The wind condition wind-speed variability used to parameterize the nominal measurement covariance was $\sigma_V=0.107~\mathrm{m\,s^{-1}}$.

Wind direction performance was also evaluated at eight different orientations, from $0^\circ$ to $315^\circ$ in $45^\circ$ increments. The maximum absolute directional error was around  $4.27^\circ$, with
circular variability of approximately $0.80^\circ$--$1.66^\circ$ across the orientations. The directional repeatability of $\sigma_\alpha=1.0 4^\circ$ was used in the nominal covariance model. The \emph{FROM} direction of the anemometer is relative to the vehicle body frame, where the sensor $0^\circ$ axis aligns with the vehicle forward direction.    

\subsection{Vehicle-Induced Flow Characterization}

A controlled propulsion experiment was conducted to determine whether the elevated sensor positioning isolates the anemometer from vehicle-induced flow. The propulsion-induced change in measured wind speed was defined as:

\begin{equation}
\Delta V
=
V_{\mathrm{prop}}-V_{\mathrm{baseline}}.
\label{eq:propwash}
\end{equation}

Across the evaluated propulsion and flow conditions, $\Delta V$ ranged approximately from $-2.13$ to
$+1.72~\mathrm{m\,s^{-1}}$. The slope was not linear with the propulsion level, showing that mounting alone does not guarantee undisturbed airborne wind observation. This observation motivates the vehicle-state-dependent confidence introduced in section IV.
\begin{figure*}[t]
    \centering

    \begin{minipage}{0.32\textwidth}
        \centering
        \includegraphics[width=\linewidth]
        {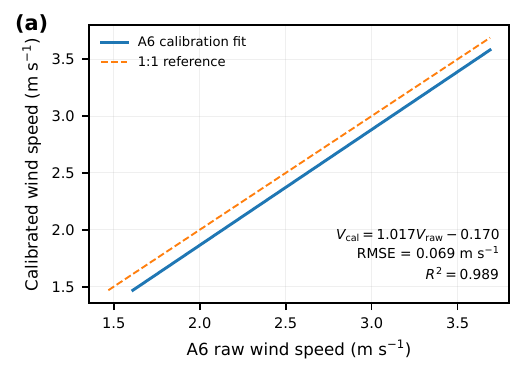}
    \end{minipage}
    \hfill
    \begin{minipage}{0.32\textwidth}
        \centering
        \includegraphics[width=\linewidth]
        {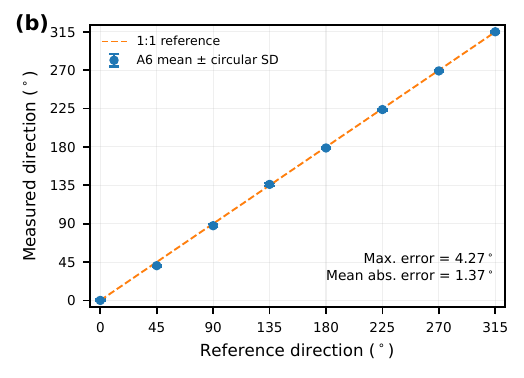}
    \end{minipage}
    \hfill
    \begin{minipage}{0.32\textwidth}
        \centering
        \includegraphics[width=\linewidth]
        {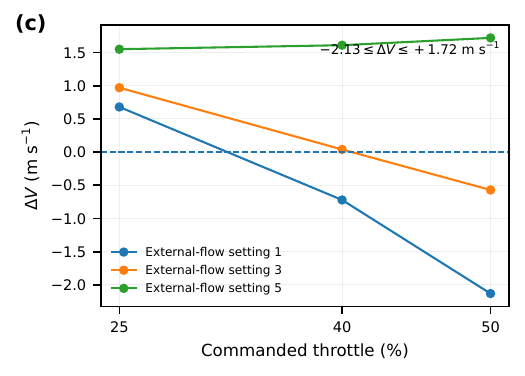}
    \end{minipage}

    \caption{
    Characterization of the representative A6 ultrasonic anemometer
    and its interaction with the multirotor platform.
    (a) Validated first-order wind-speed calibration relation over the
    experimentally evaluated range.
    (b) Wind-direction validation at eight reference orientations;
    error bars denote circular standard deviation.
    (c) Change in measured wind speed under propulsion for three
    external-flow conditions, showing that elevated mounting does not
    completely eliminate vehicle-induced aerodynamic influence.
    }
    \label{fig:a6_characterization}
\end{figure*}


\section{Vehicle-State-Aware Wind Confidence Estimation}
\label{sec:method}

\begin{figure*}[t]
    \centering
    \includegraphics[width=0.98\textwidth]
    {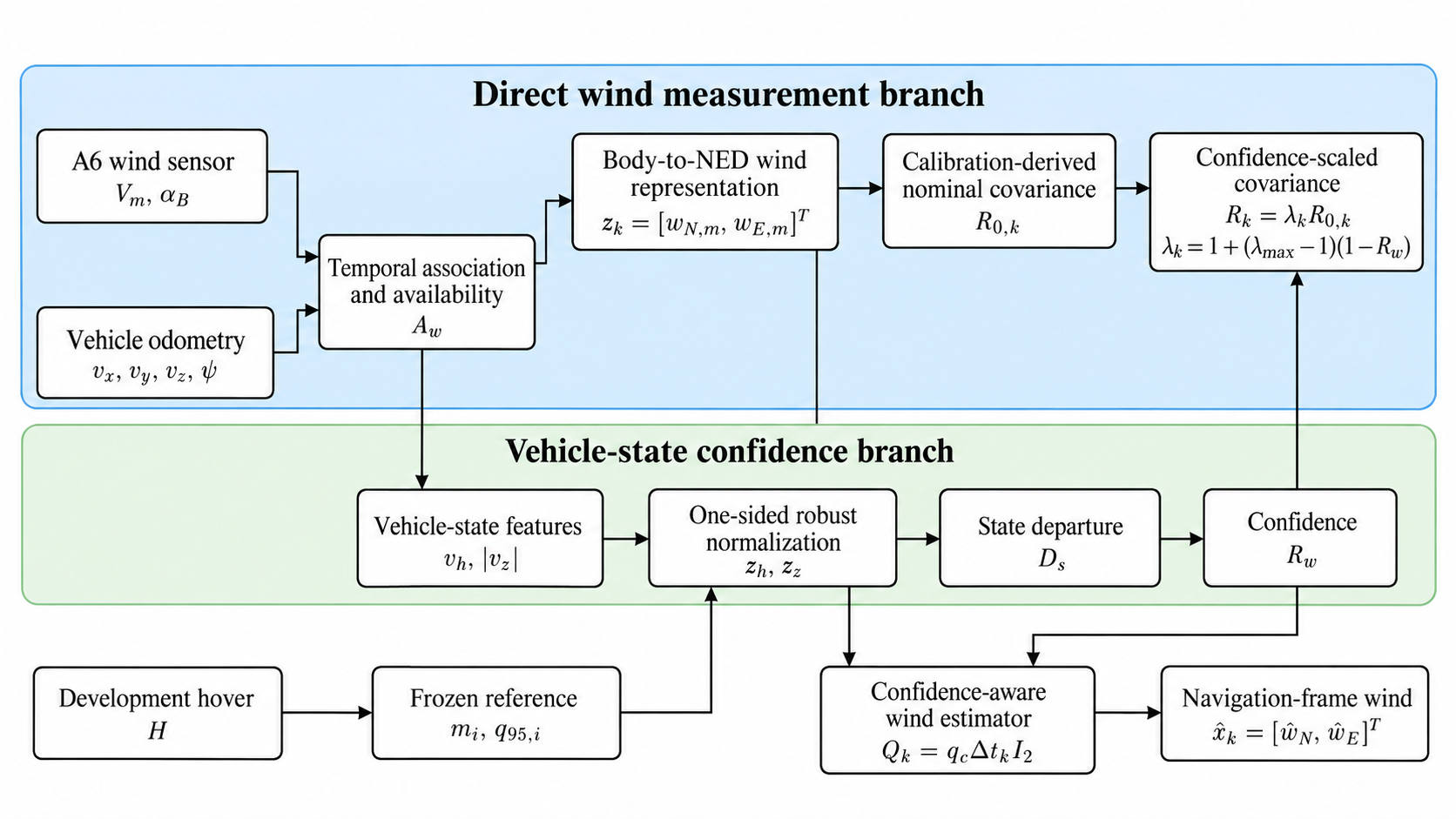}

    \caption{
    Vehicle-state-aware confidence formulation for direct multirotor
    wind sensing. Calibrated wind observations and synchronized vehicle
    state are temporally associated to determine measurement availability.
    The direct wind branch forms the navigation-frame observation and
    calibration-derived nominal covariance $R_{0,k}$, while horizontal
    and absolute vertical vehicle speeds are normalized using the frozen
    stable-hover reference to obtain the continuous confidence $R_w$.
    The resulting confidence scales the measurement covariance before
    the wind-vector estimator update.
    }
    \label{fig:confidence_workflow}
\end{figure*}

\subsection{Measurement Association and Availability}

The wind and odometry measurements are synchronized, as the data are acquired asynchronously. Each observation is associated with the nearest vehicle-state sample. The association uses a defined temporal matching tolerance, and observations that are missing or invalid/clipped for the state are considered unavailable, whereas Unreliable observations remain usable measurements acquired when the vehicle state departs from stable hover. 

where $A_w= 1$ indicates a usable synchronized observation.

\begin{equation}
A_w(t)=
\begin{cases}
1, & \text{usable synchronized wind--state observation},\\
0, & \text{otherwise}.
\end{cases}
\label{eq:availability_definition}
\end{equation}

\subsection{Vehicle-State Reconstruction}

The vehicle state is represented using the horizontal and absolute vertical speeds. The horizontal speed provides a measure of vehicle translation, while $v_{z,\mathrm{abs}}$ represents the magnitude of vertical motion irrespective of ascending or descending. Both quantities are obtained from synchronized vehicle odometry. The linear-velocity odometry components were verified to be expressed in the local NED frame.

\begin{equation}
\mathbf{s}(t)
=
\begin{bmatrix}
v_h(t) \\
v_{z,\mathrm{abs}}(t)
\end{bmatrix},
\label{eq:vehicle_state_vector}
\end{equation}

where the horizontal vehicle speed and the absolute vertical speed are defined as

\begin{equation}
v_h(t)=\sqrt{v_x^2(t)+v_y^2(t)},
\qquad
v_{z,\mathrm{abs}}(t)=|v_z(t)|.
\label{eq:vehicle_speed}
\end{equation}

These two quantities constitute the vehicle-state features $x_i \in \{v_h,v_{z,\mathrm{abs}}\}$ used in the confidence formulation.

Both features capture complementary modes of vehicle motion which influences the condition under which the wind observations are acquired. The horizontal speed $v_h$ characterizes translational motion in the horizontal plane, which affects the relative airflow experienced in the anemometer. The absolute vertical speed $v_{z,\mathrm{abs}}$ represents departure from stable hover while treating ascents and descents symmetrically.

\subsection{Stable-sensing reference state}

A stable-sensing reference is established from a validated stable hover data set, denoted by $H$. For each vehicle-state feature $x_i$, the median and 95th percentile within $H$ are defined as 

\begin{equation}
m_i
=
Q_{0.50}
\left(
x_i \mid H
\right),
\qquad
q_{95,i}
=
Q_{0.95}
\left(
x_i \mid H
\right),
\label{eq:stable_reference}
\end{equation}

where $x_i \in \{v_h,v_{z,\mathrm{abs}}\}$. The median $m_i$ provides an estimate of nominal stable hover level while $q_{95 ,i}$ sets the empirical upper reference for the motion observed during the hover. This quantile-based representation avoids a Gaussian assumption and reduces sensitivity to occasional transient values compared with normalization based on the mean and standard deviation. The reference statistics are estimated from development flight and are frozen before application to the subsequent validation data.

\subsection{One-Sided Robust State Normalization}

Each vehicle-state feature is normalized relative to its stable-sensing
reference as

\begin{equation}
z_i(t)
=
\max
\left[
0,\,
\frac{
x_i(t)-m_i
}{
q_{95,i}-m_i
}
\right],
\label{eq:one_sided_normalization}
\end{equation}

where $x_i \in \{v_h,v_{z,\mathrm{abs}}\}$. The one-sided formulation is more suitable because it penalizes only motion exceeding stable-hover median. $z_i=0$ for feature values at or below $m_i$, $z_i=1$ when $x_i=q_{95,i}$, and $z_i>1$ when feature exceeds the stable hover range. The normalized departure results are dimensionless and hence can be combined even though they are different physical scales of horizontal and vertical vehicle motion. 

\subsection{Stable-Sensing Departure}

The normalized horizontal and vertical motion features are combined into a single dimensionless measure of departure from the stable sensing reference: 

\begin{equation}
D_s(t)
=
\sqrt{
\frac{
z_h^2(t)+z_z^2(t)
}{2}
},
\label{eq:stable_sensing_departure and wind confidence}
\end{equation}

where $z_h$ and $z_z$ denote the normalized departures from horizontal and absolute vertical speed. Equal contribution is given to both normalized features using root mean square without introducing manually tuned feature weights. 

The value of $D_s=0$ reveals that both features do not exceed their stable hover median, while increasing $D_s$ indicates progressively stronger departure from the  the stable-sensing reference condition.

The state departure is then mapped to a bounded wind observation confidence score 

\begin{equation}
R_w(t)
=
\exp
\left[
-\frac{D_s^2(t)}{2}
\right],
\label{eq:wind_confidence}
\end{equation}

where $0<R_w(t)\leq1$. Values closer to one indicate the vehicle state is near stable sensing, and decreasing $R_w$ indicates a strong departure from stable sensing. The score represents the condition under which the wind observation was acquired. Importantly, $(R_w)$ is not interpreted as the probability that a wind measurement is correct. It is a dimensionless relative confidence measure derived from the vehicle-state departure from the empirically characterized stable-sensing reference. Its role is therefore to regulate measurement's influence, rather than to predict absolute wind-measurement error.     

\subsection{Confidence Interpretation and Qualitative Classes}

The continuous $R_w$ is the primary output. For visualization only, we partitioned the continuous score into Reliable for $R_w\geq0.75$, Degraded for $0.40\leq R_w<0.75$, and
Unreliable for $R_w<0.40$. These thresholds are descriptive, rather than optimized decision boundaries, and do not enter the estimator; all measurement updates use the continuous $R_w$.

Reliability represents conditions closer to stable sensing, Degraded represents moderate departure, and Unreliable represents strong departure from the reference condition. An \emph{Unavailable} state is assigned when no usable synchronized data is available ($A_w=0$).

These class thresholds are used only for visualization and descriptive labeling, whereas the continuous $R_w$ is used for subsequent confidence-aware wind processing.

\subsection{Wind-Vector Representation}

The anemometer reports the calibrated wind speed  $V_m$ and a meteorological \emph{FROM} direction $\alpha_B$ relative to the vehicle body frame. The sensor $0^\circ$ is in the same axis as the vehicle's forward axis. Since a meteorological FROM angle specifies the direction from which the airflow originates, the corresponding horizontal body-frame airflow is  
\begin{equation}
\mathbf{w}_m^{B}(t)
=
-V_m(t)
\begin{bmatrix}
\cos \alpha_B(t)\\
\sin \alpha_B(t)
\end{bmatrix},
\label{eq:wind_body}
\end{equation}

The body-frame axes are $+x$ forward and $+y$ right. The negative sign converts the reported FROM direction into the airflow vector direction.  
The body-frame vector is converted to the local NED frame using the vehicle yaw $\psi$ to express it in a common navigation frame. 

\begin{equation}
\begin{bmatrix}
w_N(t)\\
w_E(t)
\end{bmatrix}
=
\begin{bmatrix}
\cos\psi(t) & -\sin\psi(t)\\
\sin\psi(t) & \cos\psi(t)
\end{bmatrix}
\begin{bmatrix}
w_x^{B}(t)\\
w_y^{B}(t)
\end{bmatrix}.
\label{eq:wind_body_to_ned}
\end{equation}

 These navigation-frame components $w_N$ and $w_E$ are used for subsequent confidence-aware wind estimation.

\subsection{Confidence-Aware Wind Processing}

The navigation-frame wind state and corresponding measurements are represented as

\begin{equation}
\mathbf{x}_k
=
\begin{bmatrix}
w_{N,k}\\
w_{E,k}
\end{bmatrix},
\qquad
\mathbf{z}_k
=
\begin{bmatrix}
w_{N,m,k}\\
w_{E,m,k}
\end{bmatrix}.
\label{eq:filter_state_measurement}
\end{equation}

The nominal measurement covariance is derived from calibrated wind speed and direction uncertainties. In polar coordinates

\begin{equation}
\Sigma_{V\alpha}
=
\begin{bmatrix}
\sigma_V^2 & 0\\
0 & \sigma_\alpha^2
\end{bmatrix}.
\label{eq:polar_covariance}
\end{equation}

The Jacobian for the body-frame wind-vector transform is 

\begin{equation}
J_k
=
\begin{bmatrix}
-\cos\alpha_k & V_k\sin\alpha_k\\
-\sin\alpha_k & -V_k\cos\alpha_k
\end{bmatrix}.
\label{eq:wind_jacobian}
\end{equation}

Hence, the nominal measurement covariance is 

\begin{equation}
R_{0,k}
=
C_k
J_k
\Sigma_{V\alpha}
J_k^{T}
C_k^{T},
\label{eq:nominal_measurement_covariance}
\end{equation}

where $C_k$ is the body-to-NED rotation defined by the vehicle yaw.

A random-walk process model is used for short-interval wind-vector estimation

\begin{equation}
\hat{\mathbf{x}}_k^{-}
=
\hat{\mathbf{x}}_{k-1},
\qquad
P_k^{-}
=
P_{k-1}+Q_k,
\label{eq:wind_prediction}
\end{equation}

with 

\begin{equation}
Q_k
=
q_c \Delta t_k I_2,
\label{eq:process_covariance}
\end{equation}

where $q_c$ controls the nominal estimator response. 

The vehicle state confidence modifies measurement covariance through 

\begin{equation}
\lambda_k
=
1+(\lambda_{\max}-1)\bigl[1-R_w(k)\bigr],
\qquad
R_k=\lambda_kR_{0,k}.
\label{eq:adaptive_covariance}
\end{equation}

Thus, observations acquired under conditions close to the stable-sensing reference retain a measurement covariance close to  $R_{0,k}$, whereas decreasing $R_w$ increases $R_k$ and reduces the influence of the corresponding observation. 

Based on the available observation the Kalman gain and state update are

\begin{equation}
K_k
=
P_k^{-}
\left(
P_k^{-}+R_k
\right)^{-1},
\label{eq:kalman_gain}
\end{equation}

\begin{equation}
\hat{\mathbf{x}}_k
=
\hat{\mathbf{x}}_k^{-}
+
K_k
\left(
\mathbf{z}_k-\hat{\mathbf{x}}_k^{-}
\right).
\label{eq:confidence_aware_update}
\end{equation}

The covariance is updated using Joseph form, 

\begin{equation}
P_k
=
\left(I_2-K_k\right)
P_k^{-}
\left(I_2-K_k\right)^{T}
+
K_kR_kK_k^{T}.
\label{eq:joseph_update}
\end{equation}

When $A_w(k)=0$, no measurement update is performed and the predicted
state and covariance are retained.
\section{Experimental Methodology}
\label{sec:experiments}

The experiments were conducted using the sensing platform and calibrated anemometer configuration described in Section~\ref{sec:platform}. The calibration-derived uncertainty parameters defined in
Section~\ref{sec:platform} were used to construct the nominal
measurement covariance $R_{0,k}$.

\subsection{Horizontal Development Flight}

To determine the stable-sensing reference and evaluate the confidence response during translational motion,
a horizontal flight experiment was conducted. The vehicle alternated between stable-hover periods and approximately constant-speed translations between two spatial endpoints. 

The development set was constructed using this hover interval $H$, from which the reference statistics for stable sensing were obtained. The translation phases were retained only for evaluation metrics.

The evaluation interval contains 220 unique hover odometry samples and 185 unique translational samples. Vehicle-state statistics were computed from unique odometry samples to avoid repeatedly weighting the same vehicle-state sample because asynchronous topics were recorded at different rates. 

The resulting stable-hover reference values were
$m_h=0.0745~\mathrm{m\,s^{-1}}$,
$q_{95,h}=0.1798~\mathrm{m\,s^{-1}}$,
$m_z=0.0284~\mathrm{m\,s^{-1}}$, and
$q_{95,z}=0.0759~\mathrm{m\,s^{-1}}$.
These parameters were frozen before evaluation of the vertical flight.

\subsection{Vertical Validation Flight}

A vertical flight was used to evaluate the confidence formulation under vertical vehicle motion. The UAV followed a stepped altitude profile consisting of three stable hovers separated by vertical transition segments. The validation hover levels were located at  $5.36$, $7.84$, and $10.34~\mathrm{m}$.

The stable-sensing parameters obtained from the horizontal development flight were applied to this dataset. The predefined evaluation intervals contain 328 unique stable-hover odometry samples and 89 vertical transition samples. 

This experiment evaluates whether $v_{z,\mathrm{abs}}$ provides complementary information when the vehicle departs from stable hover through vertical motion, a regime not used to define the stable-sensing reference.

\subsection{Evaluation Protocol}

The proposed formulation was evaluated using feature ablation, confidence mapping-sensitivity, and confidence-aware filtering. The flight phases were defined from the experimental protocol independently of the computed confidence score, avoiding circular state labeling.   

Four feature combinations, M1--M4, were evaluated as summarized in
Table~\ref{tab:feature_ablation}.

Because an independent co-located wind reference was unavailable, ROC-AUC is used strictly to quantify separation between stable-hover and motion-disturbed acquisition regimes defined independently from the confidence score; it is not interpreted as wind-estimation accuracy. Flight-phase labels were defined from the experimental protocol before computing $R_w$, avoiding confidence-derived circular labeling. The feature ablation analysis was performed on a separate dataset. 

Secondly, sensitivity of the mapping from $D_s$ to $R_w$ was evaluated by comparing the exponential formulation with alternative inverse mapping $R_w=(1+D_s^2)^{-1}$ while the underlying state departure is unchanged.

Finally, the confidence-aware estimator was evaluated for nominal response times $\tau\in\{0.5,1,2,4\}~\mathrm{s}$ and covariance-inflation limits
$\lambda_{\max}\in\{1,2,4,8,16,32,64,128\}$.
The case $\lambda_{\max}=1$ shows the corresponding fixed-covariance baseline. Measurement influence was summarized using Kalman gain

\begin{equation}
K_{\mathrm{eff},k}
=
\frac{1}{2}\operatorname{tr}(K_k).
\label{eq:effective_gain}
\end{equation}

The analysis used synchronized wind--odometry observations for which $A_w=1$. The unavailable observations followed the prediction-only update described in Section IV. 

\section{Results}

\subsection{Vehicle-State Separation and Feature Selection}

The stable hover shows median horizontal and absolute vertical speeds of $0.081~\mathrm{m\,s^{-1}}$ and $0.031~\mathrm{m\,s^{-1}}$, respectively. During horizontal translation, the median horizontal speed increased to approximately $1.94~\mathrm{m\,s^{-1}}$, while $v_{z,\mathrm{abs}}$ remained approximately $0.032~\mathrm{m\,s^{-1}}$. Meanwhile, during the vertical transitions, the median $v_{z,\mathrm{abs}}$ increased to $0.482~\mathrm{m\,s^{-1}}$, and the median horizontal speed remained $0.111~\mathrm{m\,s^{-1}}$.

To determine whether additional attitude and angular-rate information improved separation of stable-hover and motion conditions,  Table~\ref{tab:feature_ablation} summarizes the
resulting ROC--AUC values.

\begin{table}[t]
\centering
\caption{Vehicle-state feature ablation.}
\label{tab:feature_ablation}
\begin{tabular}{lcc}
\hline
Model & Horizontal AUC & Vertical AUC \\
\hline
$M_1:\{v_h\}$                         & 0.980 & 0.639 \\
$M_2:\{v_h,v_{z,\mathrm{abs}}\}$      & 0.975 & \textbf{0.866} \\
$M_3:+\theta_{\mathrm{tilt}}$         & 0.972 & 0.851 \\
$M_4:+\omega$                         & 0.973 & 0.857 \\
\hline
\end{tabular}
\end{table}

 M2 was therefore selected as the minimum-complexity model that captures both horizontal and vertical motion. Relative to $v_h$ alone, adding  $v_{z,\mathrm{abs}}$ increased vertical AUC from 0.639 to 0.866 while retaining strong horizontal separation (AUC = 0.975). Further inclusion of tilt or angular-rate information did not improve performance consistently across the two flight regimes.

\subsection{Wind-Confidence Response}

Based on the two-feature formulation, stable hover and translational conditions were strongly separated in the horizontal development flight. For the unique odometry samples, stable hover produced a median of $R_w=0.981$ whereas the translation distribution approached zero. Approximately $90.0\%$ of the hover samples were classified as Reliable, while $91.4\%$ of translational samples were Unreliable.

The parameters obtained from the horizontal development flight were applied to the independent vertical flight without retuning. Stable hover provided a median confidence of $R_w=0.947$, whereas the vertical-transition distribution approached zero. The resulting ROC--AUC was $0.866$, with $79.9\%$ of hover samples
classified as Reliable and $69.7\%$ of vertical-transition samples classified as Unreliable.

These results show $v_h$ primarily captures horizontal translational departure, whereas $v_{z,\mathrm{abs}}$ gives complementary information required for vertical motion. 

\subsection{Held-Out Horizontal Validation}

The frozen confidence formulation was applied to an autonomous waypoint mission with repeated endpoints. The flight made repeated endpoint hovers and bidirectional translations at speeds of $1.5~\mathrm{m\,s^{-1}}$ and $2.0~\mathrm{m\,s^{-1}}$ with the frozen statistics and parameters. 

During hover, the median confidence was approximately $R_w=0.797$. In contrast, the combined translational intervals had a median confidence of $3.31\times10^{-21}$ with approximately $98.84\%$ of translation observations classified as Unreliable. The resulting hover--translation ROC--AUC was approximately $0.995$.  

The hover confidence is lower than that observed in the development flights. The median horizontal vehicle speed during the nominal hover interval is approximately $0.128~\mathrm{m\,s^{-1}}$. The frozen stable hover reference median is $0.0745~\mathrm{m\,s^{-1}}$. The nominal hover shows greater vehicle motion ,leading to lower confidence.

\begin{figure*}[t]
    \centering

    \begin{minipage}{0.98\textwidth}
        \centering
        \includegraphics[width=\linewidth]{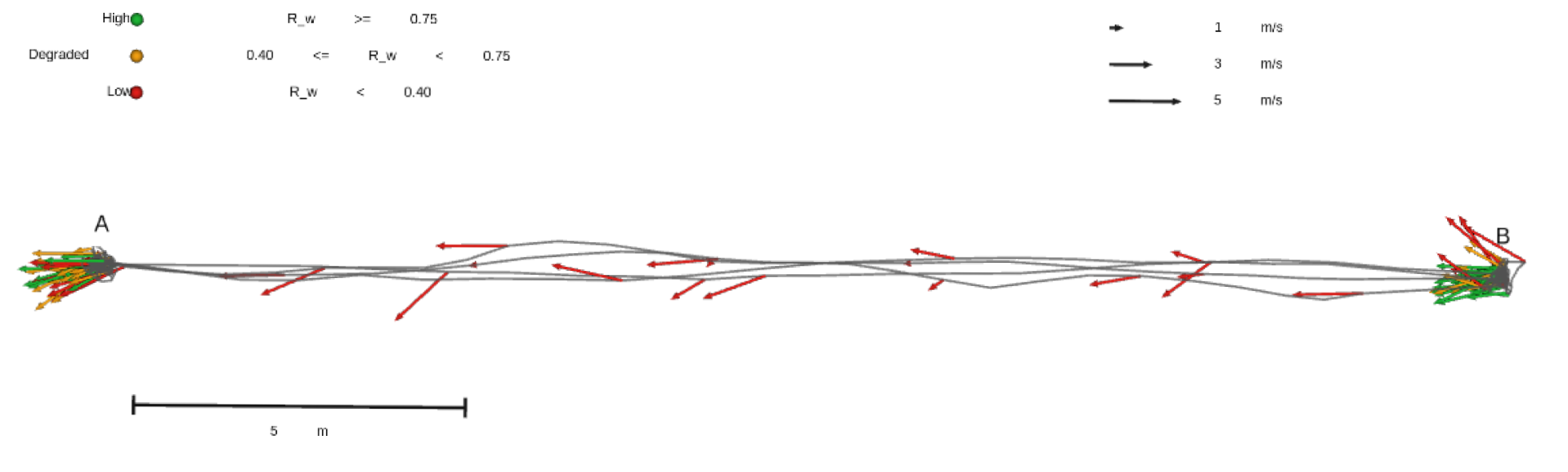}
    \end{minipage}

    \vspace{1mm}

    \begin{minipage}{0.325\textwidth}
        \centering
        \includegraphics[width=\linewidth]{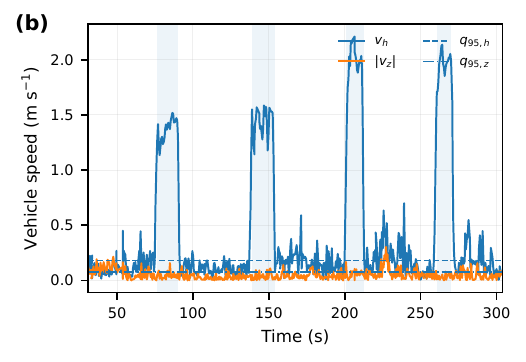}
    \end{minipage}
    \hfill
    \begin{minipage}{0.325\textwidth}
        \centering
        \includegraphics[width=\linewidth]{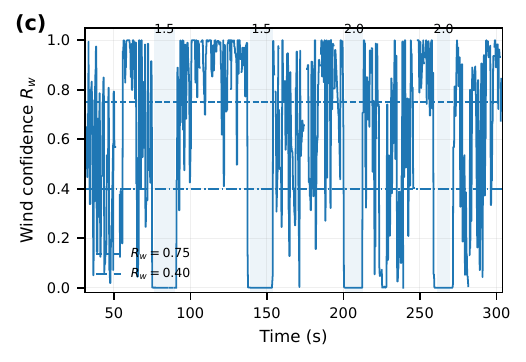}
    \end{minipage}
    \hfill
    \begin{minipage}{0.325\textwidth}
        \centering
        \includegraphics[width=\linewidth]{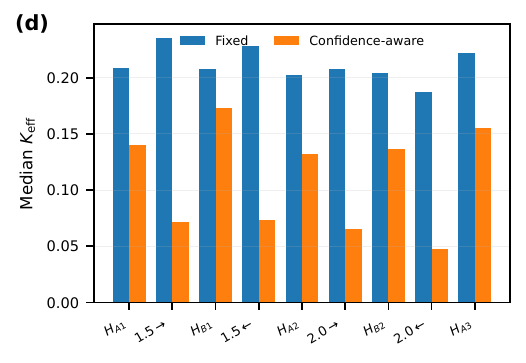}
    \end{minipage}

    \caption{
Held-out AUTO-flight validation using the frozen vehicle-state-aware
wind-confidence formulation.
(a) Spatial visualization of direct wind observations along the repeated
A--B waypoint trajectory. The gray line denotes the recorded UAV
trajectory; arrow orientation represents the measured navigation-frame
wind-vector direction, arrow length is proportional to wind speed, and
arrow color denotes confidence $R_w$. Every eighth synchronized
observation is displayed for clarity, with no spatial interpolation.
(b) Horizontal and absolute vertical vehicle-speed features relative to
the frozen stable-hover upper references; shaded regions denote
translation phases.
(c) Continuous wind-observation confidence $R_w$, with qualitative
confidence thresholds and the same independently defined translation
intervals.
(d) Median effective Kalman gain for the fixed-covariance and
confidence-aware estimators across the corresponding flight phases.
}

    \label{fig:heldout_validation}
\end{figure*}

\subsection{Confidence-Aware Estimator Response}

The adaptive covariance formulation of Section~\ref{sec:method} was used to measure the influence of continuous confidence score. The sensitivity analysis over estimator response time and covariance inflation limit resulted in the selected configuration $\tau=1~\mathrm{s}$ and $\lambda_{\max}=8$.

On the data used for parameter selection, the fixed-covariance estimator ($\lambda_{\max}=1$), produced a median effective Kalman gain of $0.2065$ during stable hover and $0.2004$ during translation, giving a translation-to-hover ratio of $0.970$. The confidence-aware estimator produced a median effective gain of $0.1885$ and $0.0743$, reducing the ratio to $0.394$.

Relative to the fixed-covariance baseline, the selected adaptive estimator retained $91.25\%$ of the median measurement influence during stable hover while reducing median measurement influence during translation by $62.91\%$. On the held-out autonomous flight, the median effective Kalman gain decreased from 0.208 to 0.151 during hover and from 0.207 to 0.064 during translation, demonstrating selective, rather than uniform attenuation of measurement influence.

The first two evaluations use unique odometry samples, while the
held-out AUTO evaluation uses available synchronized wind--state
observations.

\begin{table*}[t]
\centering
\caption{Summary of vehicle-state-aware confidence validation.}
\label{tab:main_results}
\begin{tabular}{lcccc}
\hline
Evaluation
& Hover median $R_w$
& Motion median $R_w$
& ROC--AUC
& Motion classified Unreliable (\%) \\
\hline
Horizontal development
& 0.981
& $\approx 0$
& 0.975
& 91.4 \\
Vertical validation
& 0.947
& $\approx 0$
& 0.866
& 69.7 \\
Held-out AUTO
& 0.797
& $3.31\times10^{-21}$
& 0.995
& 98.84 \\
\hline
\end{tabular}
\end{table*}

\section{Discussion and Conclusion}

\label{Interpretation}

The experiments show multirotor vehicle state provides useful context for interpreting conditions under which direct wind observation is acquired. The horizontal translation was captured by $v_h$, while $v_{z,\mathrm{abs}}$ provided complementary sensitivity to vertical translation. The two-feature formulation retains strong separation of stable hover and motion conditions without requiring additional attitude or angular rate features. 

The waypoint mission provides evidence that confidence score depends on the vehicle state rather than the commanded flight phase. Although the hover was nominal, the held-out hover intervals exhibited greater horizontal motion than the development reference, resulting in a lower confidence score. In contrast,
the translational interval has lower confidence. This is because $R_w$ represents the instantaneous sensing condition rather than a predefined label such as hover or translation.  

The confidence-aware state estimator converts the state information into selective measurement influence. Relative to the fixed-covariance baseline, the adaptive configuration retained  $91.25\%$ of the median measurement influence during stable hover while reducing the median influence during translation by $62.91\%$. This method preserves wind observations while reducing the contribution when the vehicle state departs strongly from stable sensing.   

\subsection{Limitations}

The present experiments do not include an independent, time-synchronized measurement of the undisturbed ambient wind at the UAV location. Consequently, the study establishes that vehicle state can identify departures from stable sensing conditions and regulate the influence of the corresponding observations in the estimator, but it does not claim that the resulting wind estimate is more accurate in an absolute sense. Establishing that stronger claim requires co-located or spatially registered reference airflow measurements and constitutes the principal next validation step

The stable reference statistics are based on the evaluated platform; hence, re-characterization should be done for different dynamics, sensor placements, or aerodynamic characterization.  

While the kinematics provide information about the measurement condition, they do not fully explain the aerodynamic flow at the anemometer. The approach does not correct the motor-induced flow. Future work should include an independent airborne wind reference and validation across additional platforms, flight regimes, and atmospheric conditions.

\subsection{Conclusion}

This work shows a vehicle-state-aware confidence formulation for direct wind sensing on a multirotor. Synchronized horizontal and vertical speeds were used to quantify departure from an empirically defined stable sensing reference to assign a confidence score to each available wind observation. 

Horizontal development and independent vertical validation show that the two state features provide complementary sensitivity to different regimes. Application of the frozen formulation to the waypoint flight demonstrates strong separation between low motion and translational conditions without parameter retuning. Integrating the resulting confidence with the wind estimator reduces the influence of measurements acquired during translation while largely retaining influence during stable hover. 

These results demonstrate a complementary sensing principle for aerial robots: proprioceptive state can be used not only to estimate environmental quantities directly, but also to regulate trust in direct environmental measurements. For airborne wind sensing, this enables vehicle-state-aware adaptation of measurement covariance while retaining the physical wind sensor as the primary measurement source.


\bibliographystyle{IEEEtran}
\bibliography{references_icra}

\section*{Acknowledgment}

This work was supported by the National Research Foundation under the Weather Science Research Programme (NRF-000577-01).

OpenAI ChatGPT was used for language and grammar editing of the
manuscript, equation and notation consistency checking, and the
development of Python scripts and figure layouts. The system was
used as an editing, coding, and visualization aid. All methodology,
data processing, numerical results, interpretations, and final
manuscript content were reviewed and verified by the authors.
\end{document}